\pdfoutput=1
\documentclass[pdflatex,sn-vancouver-num]{sn-jnl}

\usepackage{amsmath,amssymb}
\usepackage{graphicx}
\usepackage{array}
\usepackage{booktabs}
\usepackage{multirow}
\usepackage{algorithm}
\usepackage{algpseudocode}
\usepackage[english]{babel}

\usepackage[htt]{hyphenat}
\makeatletter
\renewcommand\paragraphfont{\reset@font\fontsize{10bp}{12bp}\bfseries\selectfont\raggedright}
\makeatother

\usepackage{bookmark}

\begin{document}

\title[SGLD vs a fixed-noise PC adaptation in canonical JEM]{Comparing SGLD and a fixed-noise predictor-corrector adaptation in canonical joint energy-based models on CIFAR-10{\renewcommand{\thefootnote}{}\footnote{This is the author's accepted manuscript. The Version of Record of this article is published in Discover Computing, and is available online at \url{https://doi.org/10.1007/s10791-026-10679-9} (open access, CC BY 4.0).}}}

\author*[1]{\fnm{Dmytro} \sur{Knopov}}\email{d.knopov@ukma.edu.ua}

\affil*[1]{\orgdiv{Faculty of Computer Sciences, Department of Mathematics}, \orgname{National University of Kyiv-Mohyla Academy}, \orgaddress{\city{Kyiv}, \country{Ukraine}}}

\abstract{
Joint Energy-Based Models (JEM) unify classification and generation within a single network and support out-of-distribution (OOD) detection.
Canonical JEM training relies on stochastic gradient Langevin dynamics (SGLD); a theoretically motivated alternative, the Predictor-Corrector (PC) sampler, has not previously undergone a systematic replication test on the canonical model.
We reproduce canonical JEM on WideResNet-28-10 without normalisation layers on two independent runs and test a fixed-noise PC adaptation -- with the degenerate annealed-noise predictor replaced by a deterministic gradient step -- across three protocols: the adapted sampler replacing SGLD throughout the full training trajectories (115--132 epochs); cold-start generation (FID); and refinement-style multi-OOD detection (AUROC).
The reconstruction reaches 92.88\% test accuracy and buffer-FID 44.46 (canonical: 92.90\% and 38.40).
We document two failure modes: catastrophic late-training divergence with the signature of the canonical outlier-buffer mechanism (all four runs), and run-dependent SVHN OOD-discrimination dynamics.
No consistent method-level advantage of the adaptation over SGLD is observed on any protocol: refinement AUROC differences stay below 0.007 across ten checkpoint--OOD pairs; seeded cold-start generation favours SGLD by about five FID points; on the training protocol a hierarchical seed-by-image bootstrap gives a 95\% confidence interval on the macro-averaged AUROC difference that contains zero, while a seed-level equivalence test with two runs per method cannot establish formal equivalence.
The training-protocol data are consistent both with equivalence and with a small directional effect.
This outcome is consistent with theory: the guarantees of the annealed-noise PC framework do not transfer to the constant-noise regime of canonical JEM.
}


\maketitle

\section{Introduction}
\label{sec:intro}

Joint Energy-Based Models (JEM), introduced by \citet{Grathwohl2020Your}, reinterpret a classifier $f_\theta(x)$ as an energy function $E_\theta(x) = -\operatorname{LSE}_y f_\theta(x)[y]$, where $\operatorname{LSE}$ denotes the log-sum-exp of the logits, and thereby enable simultaneous training of a discriminative model $p_\theta(y\mid x)$ and an implicit generative model $p_\theta(x) \propto \exp(-E_\theta(x))$.
Canonical training relies on stochastic gradient Langevin dynamics (SGLD) \citep{welling2011bayesian} for contrastive divergence; the instability of this procedure was documented by \citet{Grathwohl2020Your} themselves in Appendix~H.3, and it is precisely this instability that motivates the search for alternative samplers.
The most theoretically motivated candidate is the Predictor-Corrector (PC) sampler \citep{NEURIPS2019_3001ef25, song2021scorebased}, whose authors \citep{NEURIPS2019_3001ef25} explicitly anticipate transferring the score-based approach to energy-based models.
However, the canonical PC theory \citep{song2021scorebased} is built on a reverse-time stochastic differential equation (SDE) with an \emph{annealed} noise schedule ($\sigma_t$ is time-dependent), whereas canonical JEM uses \emph{fixed} noise $\sigma = 0.01$, so the theoretical guarantees do not transfer directly.
No systematic replication test of the PC sampler on canonical JEM has been reported in the literature.

\paragraph{Main claim.}
\emph{Within canonical JEM on CIFAR-10, a fixed-noise PC adaptation shows no detectable advantage over SGLD across three protocols -- inference-time generation quality (Fr\'echet inception distance, FID), refinement-style multi-OOD detection (area under the receiver operating characteristic curve, AUROC), and training-time sampler replacement -- in a design with two independent runs per method; on cold-start generation the adaptation is in fact detectably worse ($\Delta_{\mathrm{FID}} = +4.87 \pm 0.31$ over three paired seeds).}
Throughout the paper, \emph{PC} denotes this fixed-noise adaptation (Algorithm~\ref{alg:pc}, \S\ref{sec:pc_impl}), not the annealed-noise sampler of \citet{song2021scorebased}.
No systematic directional advantage of PC over SGLD is detected, and the hypothesis of formal equivalence at the $\pm 0.01$ AUROC margin is not established owing to the limited power of the $n = 2$ design; the data are consistent both with equivalence and with a small directional effect.
This conclusion is theoretically consistent with a limitation of the PC framework of \citet{song2021scorebased} -- the degeneration of the predictor step under fixed $\sigma$ (\S\ref{sec:why_pc_fails}).

\paragraph{Contributions.}
\begin{enumerate}
    \item \textbf{A high-fidelity canonical reconstruction} of JEM on two independent runs (\S\ref{sec:canonical}), matching the canonical accuracy ($92.88\%$ vs $92.90\%$) with an open residual FID gap ($44.46$ vs $38.40$) that may partly reflect the limited training horizon.
    \item \textbf{A characterisation of two failure modes} of canonical JEM (\S\ref{sec:failures}): catastrophic divergence with the signature of the Appendix~H.3 mechanism of \citet{Grathwohl2020Your} in all four runs in the canonical setting, and run-dependent SVHN OOD-discrimination dynamics -- to our knowledge, the first systematic two-seed documentation of this mode.
    \item \textbf{A systematic test of the PC hypothesis} (\S\ref{sec:pc_test}) on three protocols (PC replacing SGLD for the full $115$--$132$-epoch trajectories; cold-start FID and kernel inception distance (KID); refinement-style multi-OOD AUROC) with a paired and a hierarchical seed$\times$image bootstrap and a seed-level Welch two one-sided tests (TOST) procedure, yielding a verdict of ``no detectable method-level advantage under a two-run canonical evaluation'' without formally established equivalence.
\end{enumerate}

\section{Background and related work}
\label{sec:background}

\paragraph{JEM as an implicit energy-based model.}
\citet{Grathwohl2020Your} show that a classifier $f_\theta\colon \mathbb{R}^D \to \mathbb{R}^K$ defines an implicit energy-based model of the joint distribution $p_\theta(x, y) \propto \exp(f_\theta(x)[y])$ and of the marginal $p_\theta(x) \propto \exp(f_\theta^{\operatorname{LSE}}(x))$, where $f_\theta^{\operatorname{LSE}}(x) \triangleq \log\sum_y \exp(f_\theta(x)[y])$.
The maximum-likelihood gradient
\begin{equation}
    \nabla_\theta \log p_\theta(x) = \nabla_\theta f_\theta^{\operatorname{LSE}}(x) - \mathbb{E}_{x'\sim p_\theta}\bigl[\nabla_\theta f_\theta^{\operatorname{LSE}}(x')\bigr]
    \label{eq:ml_gradient}
\end{equation}
is estimated by contrastive divergence: the intractable expectation in~\eqref{eq:ml_gradient} is approximated with negative samples $x_-$ obtained by SGLD sampling from a replay buffer of size $10\,000$ with a $5\%$ reinitialisation rate \citep{Du2019Implicit}.

\paragraph{Predictor-Corrector samplers.}
A single PC iteration combines a deterministic \emph{predictor} (an integrator of the reverse-time SDE) and a stochastic \emph{corrector} (Langevin MCMC); both steps target the same density.
For the variance-exploding (VE) SDE the predictor step takes the explicit form $x'_{i} \leftarrow x_{i+1} + (\sigma_{i+1}^{2} - \sigma_{i}^{2})\,s_{\theta}(x_{i+1},\,\sigma_{i+1})$, i.e., the step length is \emph{proportional} to the difference between adjacent noise levels.
In JEM the noise is fixed ($\sigma = 0.01$), so the predictor step degenerates by construction, and in our fixed-noise adaptation (\S\ref{sec:pc_impl}, Algorithm~\ref{alg:pc}) we replace the degenerate VE predictor with a simple deterministic gradient step; this is a valid reformulation of the research question, but it takes us outside the theoretical guarantees of \citet{song2021scorebased}.
\citet{bradley2025classifier} explicitly caution that plain Langevin dynamics on a static target density without annealing -- in their words -- ``won't work in practice''.

\paragraph{Related work.}
\citet{yang2023towards} (SADA-JEM) and \citet{jiang2025your} investigate architectural modifications of canonical JEM while retaining SGLD as the sampler.
\citet{yin2025joint} (DAT) replace SGLD with a bounded deterministic projected gradient descent (PGD) sampler in a two-stage training scheme and reach FID $\approx 9$ on CIFAR-10 -- structural support for the hypothesis that the SGLD family is the limiting factor.
EBM-diffusion hybrids \citep{geng2024improving, guo2023egc, zhu2024learning, compositioncontrol2025} combine energy-based models with a native noise schedule, which restores the annealed-noise setting in which the PC guarantees apply.
\citet{nalisnick2018deep} documented the paradox of deep generative models (higher likelihood on OOD data) -- the direct precursor of Mode~2 (\S\ref{sec:fm2}).
Diffusion-based approaches to OOD detection \citep{galesso2024diffusion} form a parallel line of work outside the EBM family considered here.
\citet{bradley2025classifier} show that classifier-free guidance is a special case of the PC variant PCG, and explicitly position PCG as an analysis tool rather than a practical algorithm.
Unlike these directions, which propose new architectures or samplers and compete on generative quality, this work introduces no new method: its goal is replication -- to test, in the unchanged canonical setting of \citet{Grathwohl2020Your}, whether the theoretical advantage of PC transfers to JEM, and to document the failure modes of the canonical model.

\section{Canonical reconstruction}
\label{sec:canonical}

\paragraph{Architecture and hyperparameters.}
We implement the canonical JEM \citep{Grathwohl2020Your} with the specification cross-checked against the official \texttt{wgrathwohl/JEM} repository: WideResNet-28-10 \emph{with no normalisation layers} (\texttt{norm=None}; batch normalisation is incompatible with SGLD because it depends on batch statistics \citep{yang2023towards, yin2025joint}), LeakyReLU($0.2$) activation, \texttt{bias=True}, no dropout; Adam optimiser ($\eta = 10^{-4}$, $\lambda_{\mathrm{wd}} = 5 \cdot 10^{-4}$, \texttt{MultiStepLR} $[50, 100]$ with $\gamma = 0.3$); SGLD sampler with $K = 40$ steps, $\alpha = 1.0$, $\sigma = 0.01$, clamping $x_k \in [-1, 1]$; replay buffer of size $10\,000$ with a $5\%$ reinitialisation rate; input augmentation \texttt{Pad4 + RandomCrop32 + HFlip + Normalize} plus Gaussian noise $\sigma_{\mathrm{in}} = 0.03$; batch size $64$.
The $K = 40$ stabiliser (Appendix~H.3 of \citep{Grathwohl2020Your}) is applied from the first epoch onward, which corresponds to a pure canonical reconstruction without the two-phase $20 \to 40$ switch.
The canonical training procedure is formalised in Algorithm~\ref{alg:jem_training}.

\begin{algorithm}[htbp]
\caption{Canonical JEM training with the SGLD sampler and replay buffer}
\label{alg:jem_training}
\begin{algorithmic}[1]
\Require Dataset $\{(x_i, y_i)\}$; parameters $\theta$; buffer $\mathcal{B}$ of size $N_{\mathcal{B}} = 10\,000$; $K, \alpha, \sigma, \rho, \sigma_{\mathrm{in}}$; Adam $(\eta, \lambda)$
\State Initialise $\theta$ (He init); $\mathcal{B} \sim \mathcal{U}(-1,1)$
\For{each minibatch iteration}
    \State $(x_+, y) \gets$ augmented batch; $x_+ \gets \operatorname{clamp}(x_+ + \mathcal{N}(0,\sigma_{\mathrm{in}}^2 I),\,-1,\,1)$
    \State $x_- \gets$ draw from $\mathcal{B}$ with per-sample reinitialisation probability $\rho$
    \For{$k = 1, \dots, K$}
        \State $x_- \gets \operatorname{clamp}\bigl(x_- + \alpha\,\nabla_{x_-} f_\theta^{\operatorname{LSE}}(x_-) + \sigma\,\varepsilon_k,\,-1,\,1\bigr),\;\varepsilon_k \sim \mathcal{N}(0, I)$
    \EndFor
    \State $\mathcal{L}(\theta) \gets \mathcal{L}_{\mathrm{CE}}(x_+, y;\,\theta) - \bigl[f_\theta^{\operatorname{LSE}}(x_+) - f_\theta^{\operatorname{LSE}}(x_-)\bigr]$
    \State $\theta \gets \mathrm{Adam}_{\eta,\,\lambda}(\theta, \nabla_\theta \mathcal{L})$; return $x_-$ to $\mathcal{B}$
\EndFor
\end{algorithmic}
\end{algorithm}

\paragraph{Canonical reproduction on two runs.}
The reconstruction was performed on two independent runs differing only in the value of \texttt{SEED} ($42$ and $123$); both remained stable for $115$ and $120$ epochs respectively before collapsing with the Appendix~H.3 signature.
The divergence epochs quoted throughout are the mass-abort epochs at which training terminated: epoch~116 for \texttt{SEED}~=~42, whose collapse and abort coincide, and epoch~122 for \texttt{SEED}~=~123, whose collapse epoch~121 was itself still written to the log (\S\ref{sec:fm1}).
Checkpoints are saved every $5$ epochs; the comparison against the canonical reference values (Table~\ref{tab:canonical}) uses the \emph{last-stable} protocol -- the last saved stable checkpoint before divergence -- while the fair cross-method comparison of PC vs SGLD in \S\ref{sec:pc_test} uses the \emph{margin-10} protocol, i.e., the checkpoint $\approx 10$ epochs before each run's own divergence point.
The last-stable offsets from divergence are asymmetric because of checkpointing discreteness: $\approx 11$ epochs for \texttt{SEED}~=~42 (ep~105 vs divergence at ep~116) and $\approx 2$ epochs for \texttt{SEED}~=~123 (ep~120 vs ep~122); the peak accuracy of \texttt{SEED}~=~123 is reached at epoch~101 ($92.98\%$), while the remaining metrics in Table~\ref{tab:canonical} are evaluated at the last-stable point for a homogeneous within-run comparison.
The numerical discrepancies for \texttt{SEED}~=~123 between Table~\ref{tab:canonical} (last-stable, ep~120) and Table~\ref{tab:pc_training_eval} (margin-10, ep~110) follow from this separation of protocols, not from any inconsistency in the data.

\begin{table}[htbp]
    \centering
    \caption{Canonical reconstruction on two runs at the \emph{last-stable} checkpoints, compared with the reference values from \citet{Grathwohl2020Your}.
    The accuracy of both runs lies within $\pm 0.5$~percentage points (pp) of the canonical reference; buffer-FID exceeds the canonical $38.40$ by $\approx 6$ points, which may partly reflect the limited training horizon (our runs diverge at epochs~116--122, whereas the canonical trajectory continued to $150$ epochs)}
    \label{tab:canonical}
    \small
    \begin{tabular}{lcccc}
        \toprule
        \textbf{Metric} & \textbf{\texttt{SEED}~=~42} & \textbf{\texttt{SEED}~=~123} & \textbf{Grathwohl} & \textbf{Deviation} \\
        & (ep~105) & (ep~120) & 2020 & (best of two) \\
        \midrule
        Test accuracy & $92.88\%$ & $92.45\%$ & $92.90\%$ & $-0.02$~pp \\
        Buffer-FID & $44.46$ & $45.02$ & $38.40$ & $+6.06$ \\
        AUROC vs SVHN (static) & $0.568$ & $0.608$ & $0.670$ & $-0.062$ \\
        Divergence epoch & ep~116 & ep~122 & --- & --- \\
        \bottomrule
    \end{tabular}
\end{table}

\begin{figure}[htbp]
    \centering
    \includegraphics[width=0.85\linewidth]{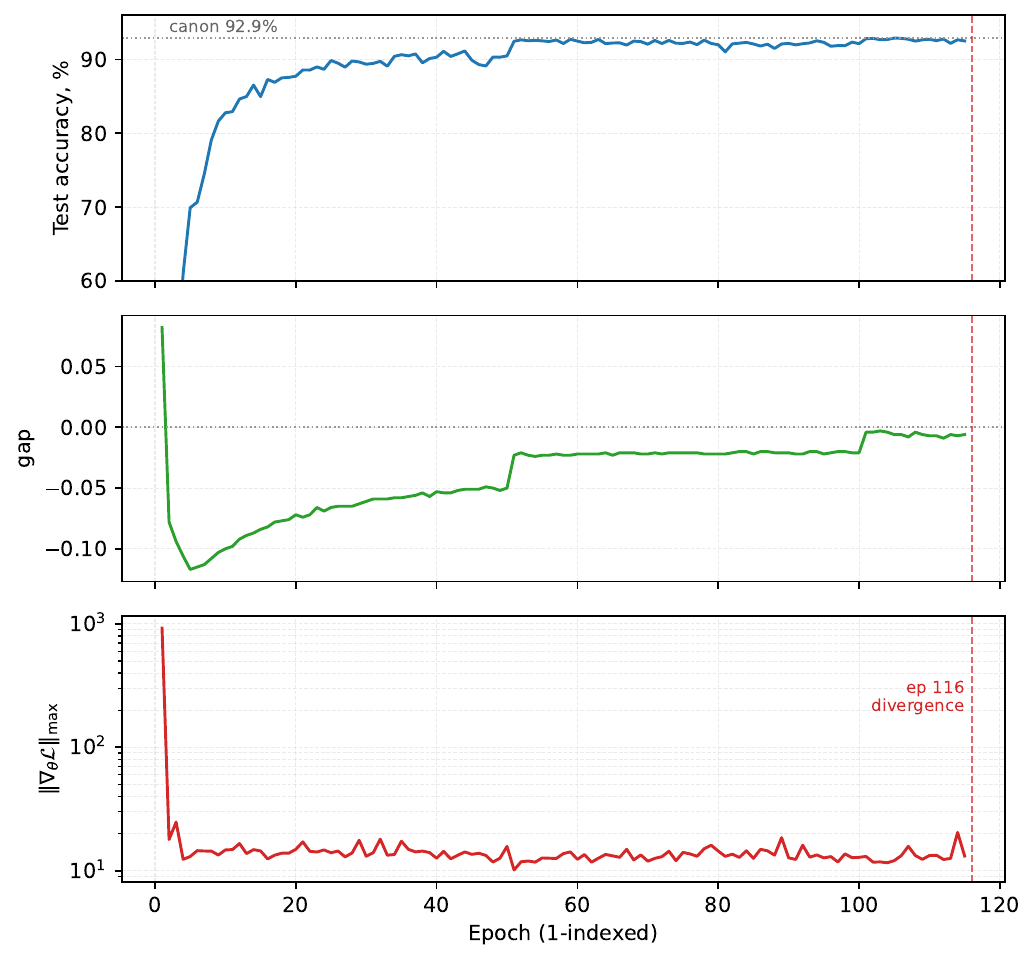}
    \caption{Training dynamics of the canonical SGLD run \texttt{SEED}~=~42 (accuracy, $\mathrm{gap} \triangleq f_\theta^{\operatorname{LSE}}(x_+) - f_\theta^{\operatorname{LSE}}(x_-)$, and the maximum model gradient norm).
    Classification accuracy rises throughout the stable phase with small epoch-to-epoch fluctuation; after the first-epoch initialisation transient the $\mathrm{gap}$ remains consistently negative up to the divergence epoch (dashed line, epoch~116 -- the mass-abort epoch, for which no aggregate metrics were logged; \S\ref{sec:fm1})}
    \label{fig:training_curves}
\end{figure}

\begin{figure}[htbp]
    \centering
    \includegraphics[width=0.95\linewidth]{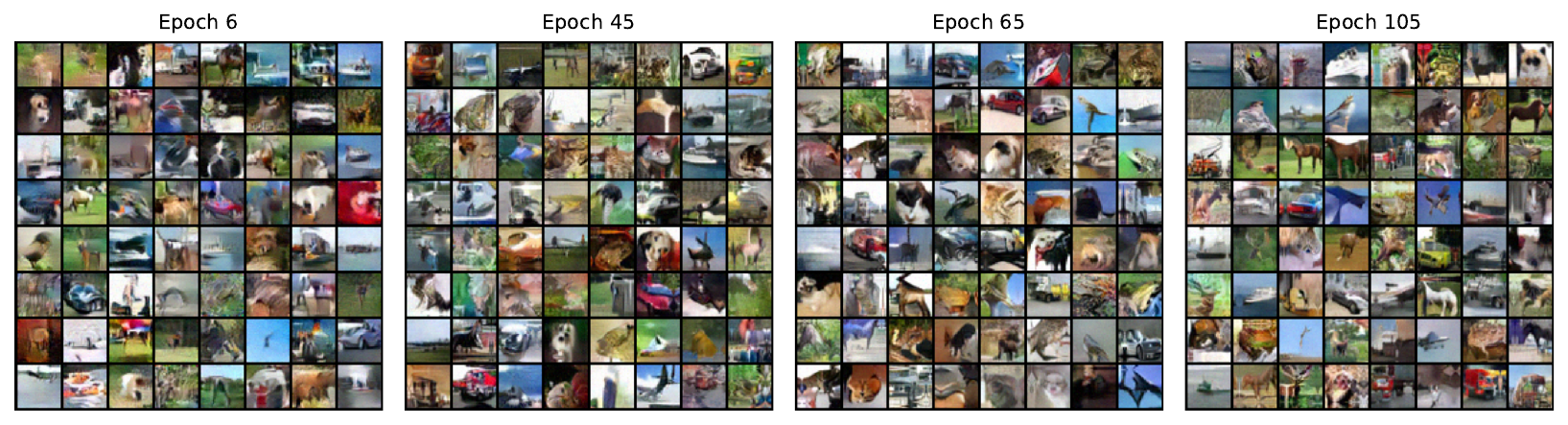}
    \caption{Replay-buffer samples across training in the canonical SGLD run (\texttt{SEED}~=~42).
    By epoch~6 the buffer already contains coherent, recognisable samples from diverse classes, which persist through epochs~45, 65 and~105, consistent with the buffer-FID improving from $\approx 49$ to $44$--$45$ at the last-stable checkpoints (Table~\ref{tab:canonical})}
    \label{fig:buffer_samples}
\end{figure}

The systematic $\sim 6$-point buffer-FID gap in both runs may partly reflect the limited training horizon (our reconstructions diverge earlier than the canonical $150$ epochs), although this explanation has not been tested directly; the replay buffer nonetheless holds coherent, recognisable samples from early in training onward (Figure~\ref{fig:buffer_samples}).
Classification accuracy is reproduced to within $0.02$~pp of the canonical value, indicating the correctness of the architecture and procedure.

\section{Failure modes of canonical JEM}
\label{sec:failures}

The canonical reconstruction produced a working model but revealed two qualitatively different failure modes under strict adherence to the canonical hyperparameters.
One of them was anticipated by \citet{Grathwohl2020Your}; the second, to our knowledge, is not documented in \citep{Grathwohl2020Your, yang2023towards, yin2025joint}.

\subsection{Mode 1: catastrophic divergence with the Appendix H.3 signature}
\label{sec:fm1}

Both SGLD runs trained stably throughout the whole expected horizon (test accuracy above $90\%$ from roughly epoch~50 onward, zero abort iterations) and then underwent catastrophic divergence: \texttt{SEED}~=~42 at epoch~116, \texttt{SEED}~=~123 at epoch~122.
The between-run range is $6$ epochs ($\sim 5\%$ of the training horizon).
The observed event signature is identical: within a single epoch the loss grows by $3$--$4$ orders of magnitude, the gap inverts from $\approx -0.01$ in the final stable epochs to $+36$--$+38$, test accuracy collapses to $12$--$24\%$ (in the two runs whose collapse epoch was written to the log), and the model gradient norm jumps by $3$--$4$ orders of magnitude within a single batch.
Figure~\ref{fig:training_curves} traces the \texttt{SEED}~=~42 run up to its mass-abort epoch, for which no aggregate metrics were logged; the collapse-epoch values quoted here are therefore those of the two runs whose pre-abort epoch was still written to the log.
Most batches in that epoch (for run \texttt{SEED}~=~42, $551$ of $781$) trigger a sharp increase in the loss, after which training terminates in a mass emergency abort.

The trigger -- a gradient spike on a single batch -- is consistent with the instability mechanism that \citet{Grathwohl2020Your} describe in Appendix~H.3: drawing a sample with anomalously high energy from the replay buffer produces a gradient orders of magnitude above the stable level and pushes the model towards divergence.
In the two runs whose final pre-abort epoch was still written to the log, the collapse is confined to the negative-sample branch: $f_\theta^{\operatorname{LSE}}(x_-)$ collapses to $-37.4$ and $-38.2$ from $\approx +0.3$ in the preceding stable epochs, whereas $f_\theta^{\operatorname{LSE}}(x_+)$ moves only to $-0.6$ and $-1.0$ -- a shift of order one against a shift of order forty -- and training accuracy is still $\geq 99.5\%$, which is consistent with a replay-buffer origin of the trigger; the per-batch records needed to identify the specific triggering sample were not retained, so we report an association, not a demonstrated cause.
These epoch aggregates moreover exclude the skipped batches themselves, so they are lower bounds on the anomaly.
The cascade signature -- \emph{a gradient spike on an outlier sample $\to$ gap inversion $\to$ accuracy collapse} -- reproduces in all four of our runs in the canonical setting (two SGLD- and two PC-trained, \S\ref{sec:pc_training}).
In all four runs the divergence occurs with the learning rate already at the floor of the canonical schedule.
The canonical optimiser is Adam at an initial rate of $10^{-4}$, with a multi-step schedule that multiplies the rate by $0.3$ at epochs~$50$ and~$100$; no further decay is scheduled, so from epoch~$100$ onward the rate remains at the floor value of $9 \cdot 10^{-6}$, and every divergence event reported here (epochs~$115$--$132$ across the four runs) falls inside that final segment.
This does not by itself establish that the event is learning-rate independent, but it does show that the scheduled decays alone do not prevent it.
Standard interventions offer little remedy: \citet{Grathwohl2020Your} report in the same appendix that gradient clipping, energy clipping, and discarding examples with atypical energies failed to resolve the instability, while decreasing the learning rate and increasing the number of SGLD steps slow training without removing the failure mode; an adaptive learning-rate policy is untested in this setting and, given the floor-LR observation above, would plausibly delay rather than prevent the event.

Our canonical configuration follows that report and applies neither gradient clipping nor a gradient-norm abort criterion, so no clipping threshold was in force at any point during the four runs and no such threshold could have been triggered; the gradient norms reported here are raw values.
The guards that were active are loss-based, and all three write to the training log: a batch whose loss exceeds $10^{4}$ in absolute value is skipped before the backward pass and therefore contributes no gradient; an epoch in which more than half of the batches are skipped in this way terminates the run; and a non-finite loss terminates it immediately.
Independently of these guards, the global $\ell_2$ norm of the parameter gradient was computed after every backward pass and its per-epoch maximum recorded for every epoch.
Excluding the first epoch of each run, where the initialisation transient reaches $8.6 \cdot 10^{2}$ to $1.1 \cdot 10^{3}$, this per-epoch maximum never exceeds $31.8$ in any run; its median over the $479$ non-collapse logged epochs is $13.4$.
In the two runs whose pre-abort epoch was still written to the log it reaches $7.50 \cdot 10^{4}$ (\texttt{SEED}~=~123, epoch~121) and $8.41 \cdot 10^{4}$ (PC \texttt{SEED}~=~42, epoch~131).
The anomaly is thus separated from the ordinary fluctuations of the stable phase by three to four orders of magnitude, and, because the loss-panic guard removes the most extreme batches before they reach the backward pass, these maxima remain lower bounds.
The definitive causal tests -- per-batch diagnostics of the update sequence beyond aggregate gradient norms, and an intervention that removes, clips, or replaces the suspected buffer outlier at the trigger point -- require re-running training with forensic logging and are left for future work.

\subsection{Mode 2: run-dependent energy-landscape dynamics on SVHN}
\label{sec:fm2}

The second mode manifests not as a sudden breakdown but as irregular dynamics in canonical JEM's ability to discriminate in-distribution from out-of-distribution data using the energy score during the stable phase.
We measured the static score $s(x) = -\operatorname{LSE} f_\theta(x)$ at seven checkpoints of each of the two SGLD runs (Table~\ref{tab:ood_trajectory}).
This quantity is the energy score of \citet{liu2020energy} at temperature $T = 1$, taken here with the opposite sign convention so that larger values indicate greater atypicality.
Within the JEM parameterisation it further equals $-\log p_\theta(x)$ up to the log partition function -- an identification their detector deliberately avoids, since it is defined without reference to a normalised density.

\begin{table}[htbp]
    \centering
    \caption{Dynamics of the static OOD metric (CIFAR-10 vs SVHN) on the two canonical SGLD runs.
    The landscape state is determined by the mean energies: ``inverted'' $\Leftrightarrow$ $\overline{E}_{\mathrm{SVHN}} < \overline{E}_{\mathrm{CIFAR}}$ (column $E_{\mathrm{CIFAR}}$ vs $E_{\mathrm{SVHN}}$), ``healthy'' -- the opposite.
    AUROC is an accompanying rank-based metric, consistent with the state at $13$ of the $14$ checkpoints; the exception is \texttt{SEED}~=~123, ep~85 (AUROC $= 0.522$ with $\overline{E}_{\mathrm{SVHN}} < \overline{E}_{\mathrm{CIFAR}}$), where the ``inverted'' label is kept according to the primary energy criterion.
    Bold marks the peak-healthy checkpoint of each run, that is, the checkpoint reported for that run in Table~\ref{tab:canonical}}
    \label{tab:ood_trajectory}
    \small
    \setlength{\tabcolsep}{5pt}
    \begin{tabular}{ccccl}
        \toprule
        \textbf{Epoch} & \textbf{Accuracy} & \textbf{AUROC vs SVHN} & $\boldsymbol{E_{\mathrm{CIFAR}}}$ \textbf{vs} $\boldsymbol{E_{\mathrm{SVHN}}}$ & \textbf{Landscape state} \\
        \midrule
        \multicolumn{5}{l}{\emph{Run \texttt{SEED}~=~42 (divergence at ep~116)}} \\
        25  & $89.84\%$ & $0.486$ & $-0.519$ vs $-0.557$ & inverted \\
        45  & $89.88\%$ & $0.424$ & $-0.449$ vs $-0.545$ & inverted \\
        65  & $92.21\%$ & $0.558$ & $-0.359$ vs $-0.332$ & healthy \\
        85  & $92.08\%$ & $0.418$ & $-0.254$ vs $-0.369$ & inverted \\
        90  & $92.16\%$ & $0.499$ & $-0.310$ vs $-0.331$ & inverted \\
        \textbf{105} & $92.88\%$ & $0.568$ & $-0.259$ vs $-0.205$ & \textbf{healthy (peak)} \\
        115 & $92.49\%$ & $0.497$ & $+0.026$ vs $-0.214$ & inverted \\
        \midrule
        \multicolumn{5}{l}{\emph{Run \texttt{SEED}~=~123 (divergence at ep~122)}} \\
        25  & $89.05\%$ & $0.348$ & $-0.471$ vs $-0.648$ & inverted \\
        45  & $90.89\%$ & $0.338$ & $-0.314$ vs $-0.513$ & inverted \\
        65  & $92.55\%$ & $0.544$ & $-0.300$ vs $-0.283$ & healthy \\
        85  & $92.23\%$ & $0.522$ & $-0.230$ vs $-0.275$ & inverted \\
        100 & $92.44\%$ & $0.556$ & $-0.328$ vs $-0.280$ & healthy \\
        105 & $92.43\%$ & $0.580$ & $-0.147$ vs $-0.062$ & healthy \\
        \textbf{120} & $92.45\%$ & $0.608$ & $-0.203$ vs $-0.081$ & \textbf{healthy (peak)} \\
        \bottomrule
    \end{tabular}
\end{table}

\begin{figure}[htbp]
    \centering
    \includegraphics[width=0.85\linewidth]{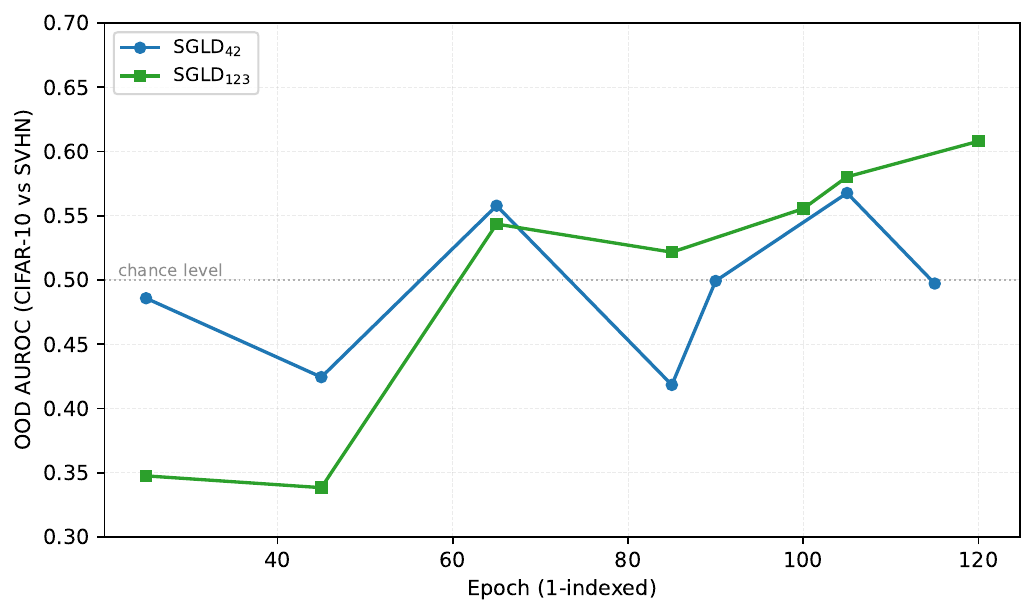}
    \caption{Trajectory of the static OOD AUROC (CIFAR-10 vs SVHN) over training for the two canonical SGLD runs.
    Both runs start in the inverted regime ($\text{AUROC} < 0.5$) and leave it at approximately epoch~65; \texttt{SEED}~=~42 subsequently oscillates (peak $0.568$ at ep~105, re-inversion at ep~115), while \texttt{SEED}~=~123 improves steadily ($0.556 \to 0.580 \to 0.608$).
    The two runs share architecture and hyperparameters and differ only in \texttt{SEED}, so the divergent OOD dynamics are a run-level effect (Mode~2)}
    \label{fig:ood_trajectory}
\end{figure}

Both runs start in the inverted regime at early epochs ($\text{AUROC} < 0.5$) and leave it at approximately epoch~65, but the subsequent dynamics differ substantially: \texttt{SEED}~=~42 oscillates between the inverted and healthy states (peak $0.568$ at ep~105, returning to inversion at ep~115), whereas \texttt{SEED}~=~123 improves steadily ($0.556 \to 0.580 \to 0.608$ over ep~100--120; Figure~\ref{fig:ood_trajectory}).
The between-run range at the headline checkpoints is $0.040$ AUROC, i.e., $\sim 8\%$ of the working range $[0.5, 1.0]$.
The canonical reading of JEM as a model that ``consistently assigns higher likelihoods\ldots'' \citep[Section~5.3.1]{Grathwohl2020Your} rests on single-run estimates and, according to our data, turns out to be a seed-dependent property rather than a stable attribute of the JEM architecture.
Replication on at least two \texttt{SEED}s is a minimal requirement for assessing the robustness of canonical JEM as an OOD detector.

\section{Empirical testing of the Predictor-Corrector hypothesis}
\label{sec:pc_test}

The documented failure modes motivate testing alternative samplers.
We systematically test the PC adaptation on three complementary protocols: \textbf{Protocol~1} -- PC instead of SGLD throughout the full training trajectories, $115$--$132$ epochs to divergence (\S\ref{sec:pc_training}); \textbf{Protocol~2} -- PC vs SGLD on cold-start generation with FID/KID evaluation over three paired seeds (\S\ref{sec:pc_fresh_fid}); \textbf{Protocol~3} -- PC vs SGLD as a refinement-style OOD score on five datasets (\S\ref{sec:pc_refinement_ood}).
All samplers receive an identical budget of gradient evaluations ($40$ per batch in Protocol~1, $20$ in Protocol~3, $1000$ in Protocol~2).

\subsection{A fixed-noise PC adaptation for canonical JEM}
\label{sec:pc_impl}

Our PC scheme (Algorithm~\ref{alg:pc}) adapts the formulation of \citet{song2021scorebased} to the constant-noise regime of canonical JEM, taking $\nabla_x f_\theta^{\operatorname{LSE}}(x)$ as the score estimate.

\begin{algorithm}[htbp]
\caption{Fixed-noise Predictor-Corrector adaptation for JEM}
\label{alg:pc}
\begin{algorithmic}[1]
\Require Initial point $x_0$; iterations $K$; corrector steps $K_c$; step size $\alpha$; noise $\sigma$
\State $x \gets x_0$
\For{$k = 1, \dots, K$}
    \State $x \gets \operatorname{clamp}(x + \alpha\,\nabla_x f_\theta^{\operatorname{LSE}}(x),\;-1,\;1)$ \Comment{Predictor -- deterministic}
    \For{$j = 1, \dots, K_c$}
        \State $x \gets \operatorname{clamp}(x + \alpha\,\nabla_x f_\theta^{\operatorname{LSE}}(x) + \sigma\,\varepsilon,\;-1,\;1),\;\varepsilon \sim \mathcal{N}(0, I)$ \Comment{Corrector -- stochastic}
    \EndFor
\EndFor
\end{algorithmic}
\end{algorithm}

\paragraph{Deliberate deviations.}
The adaptation differs from \citet{song2021scorebased} in two respects: (i)~a fixed corrector step size instead of the adaptive signal-to-noise ratio -- for consistency with canonical SGLD; (ii)~constant noise $\sigma = 0.01$ instead of an annealed schedule $\sigma_t$ -- because canonical JEM has none.
These deviations take our adaptation outside the theoretical guarantees of \citet{song2021scorebased}, which we discuss in \S\ref{sec:why_pc_fails}, but they enable a direct comparison with canonical SGLD at an equal computational budget.

\subsection{Protocol 1: PC as the training-time sampler}
\label{sec:pc_training}

Protocol~1 investigates the setting in which PC replaces SGLD in the contrastive-divergence procedure for the full training trajectory ($115$ and $132$ epochs to divergence in the two PC runs).
All hyperparameters except the sampler itself coincide with the canonical variant; PC is configured with $K_c = 1$ and $K_{\text{outer}} = 20$ ($40$ gradient evaluations per batch in total, equal to the canonical $K = 40$).
Two independent PC runs with \texttt{SEED}~$\in\{42, 123\}$ were performed to assess the between-run range; both reached comparable peak accuracies ($92.81\%$ and $92.85\%$) and underwent catastrophic divergence with the same Appendix~H.3 signature: PC \texttt{SEED}~=~42 at epoch~132, PC \texttt{SEED}~=~123 at epoch~115.

To compare the PC- and SGLD-trained models, we evaluated all four runs in the canonical setting by the static OOD score $s(x) = -\operatorname{LSE} f_\theta(x)$ (which does not depend on the sampler) on five datasets, at each run's own margin-10 checkpoint (Table~\ref{tab:pc_training_eval}).
The five sets combine the two publicly redistributable benchmarks of the original JEM evaluation -- SVHN and CIFAR-100 \citep{Grathwohl2020Your} -- with three standard CIFAR-10 OOD benchmarks from the detection literature: DTD/Textures, LSUN-R, and iSUN \citep{liu2020energy}, covering far-OOD, near-OOD, texture, and scene shifts.

\begin{table}[htbp]
    \centering
    \caption{Evaluation of the four runs in the canonical setting at the margin-10 checkpoints: per-dataset static OOD AUROC and method means.
    Checkpoints are $\approx 10$ epochs before each run's own divergence point.
    Bold marks the best result in each row.
    Here and in Tables~\ref{tab:multi_ood} and~\ref{tab:summary}, where differences between the samplers are of order $0.005$ or smaller, AUROC is quoted to four decimal places; Tables~\ref{tab:canonical} and~\ref{tab:ood_trajectory}, which compare against a published reference value and track a trajectory with swings of order $0.1$, use three decimal places}
    \label{tab:pc_training_eval}
    \footnotesize
    \setlength{\tabcolsep}{3pt}
    \begin{tabular}{l cc cc}
        \toprule
        & \multicolumn{2}{c}{\textbf{Canonical SGLD}} & \multicolumn{2}{c}{\textbf{PC-trained}} \\
        \cmidrule(lr){2-3}\cmidrule(lr){4-5}
        \textbf{OOD dataset} & \texttt{SEED}~=~42 & \texttt{SEED}~=~123 & \texttt{SEED}~=~42 & \texttt{SEED}~=~123 \\
        & (ep~105) & (ep~110) & (ep~120) & (ep~105) \\
        \midrule
        SVHN         & $0.5676$ & $0.6056$ & $0.6045$ & $\mathbf{0.6237}$ \\
        CIFAR-100    & $0.6396$ & $0.6244$ & $\mathbf{0.6418}$ & $0.6391$ \\
        DTD/Textures & $0.7250$ & $0.7349$ & $0.7374$ & $\mathbf{0.7549}$ \\
        LSUN-R       & $\mathbf{0.7805}$ & $0.7688$ & $0.7511$ & $0.7737$ \\
        iSUN         & $0.7863$ & $\mathbf{0.7899}$ & $0.7668$ & $0.7859$ \\
        \midrule
        \textbf{macro-5 OOD AUROC} & $0.6998$ & $0.7047$ & $0.7003$ & $\mathbf{0.7155}$ \\
        \midrule
        \textbf{Method mean} & \multicolumn{2}{c}{$\overline{\text{macro-5}}_{\mathrm{SGLD}} = 0.7022$} & \multicolumn{2}{c}{$\overline{\text{macro-5}}_{\mathrm{PC}} = 0.7079$} \\
        \textbf{Between-run range} & \multicolumn{2}{c}{$0.0049$} & \multicolumn{2}{c}{$0.0152$} \\
        \bottomrule
    \end{tabular}
\end{table}

\begin{figure}[htbp]
    \centering
    \includegraphics[width=0.9\linewidth]{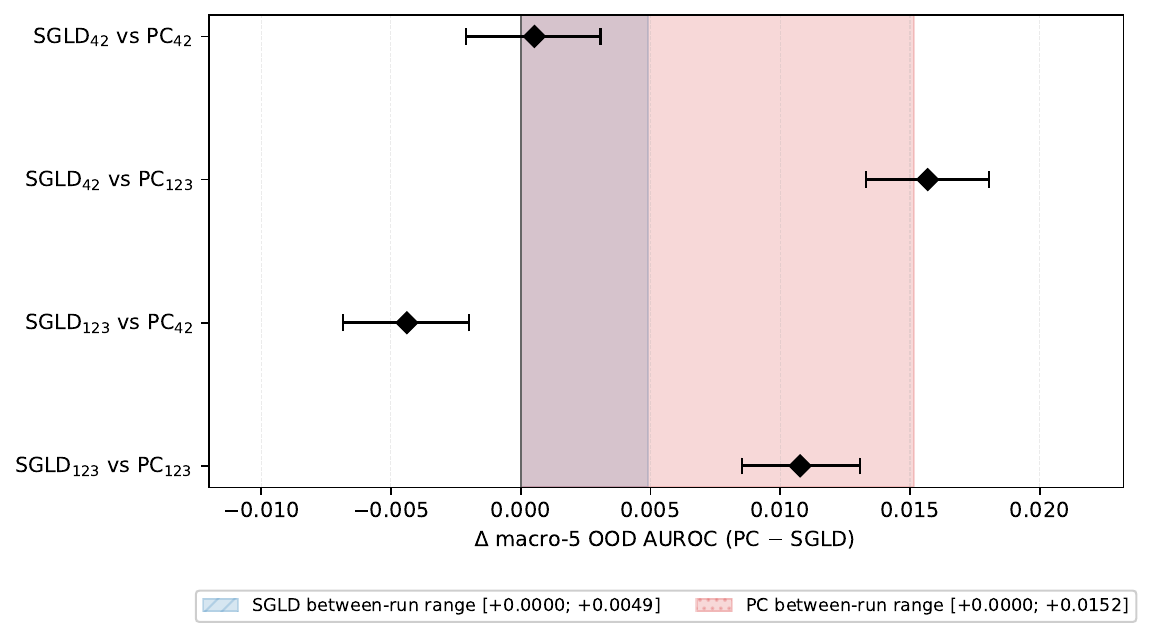}
    \caption{Training Protocol~1: cross-method $\Delta\text{macro-5 AUROC}$ at the margin-10 checkpoints (four cross-method pairs) with paired-bootstrap $95\%$ CIs ($B = 3000$; each iteration jointly resamples one shared in-distribution index vector and independent per-dataset out-distribution vectors, computing both methods' macro-5 on the same indices); the shaded bands show the within-method ranges -- $0.0049$ for SGLD and $0.0152$ for PC.
    All four $\Delta$ point estimates lie between $-0.0044$ and $+0.0157$, and in two of the four pairs $|\Delta|$ does not exceed the SGLD within-method range.
    The per-image CIs quantify scoring noise at fixed checkpoints only; method-level uncertainty is assessed by the hierarchical analysis of \S\ref{sec:pc_summary}}
    \label{fig:pc_vs_sgld}
\end{figure}

The method-level shift of the mean macro-5 AUROC is $\Delta = +0.0056$ in favour of PC, but this is smaller than the PC between-run range ($0.0152$, three times wider than that of SGLD).
In the paired cross-method comparisons the direction of the effect is heterogeneous (Figure~\ref{fig:pc_vs_sgld}): at the per-image level and at fixed checkpoints, two of the four pairings significantly favour PC, one favours SGLD, and one shows no detectable difference.
These per-pair intervals quantify scoring noise alone; the method-level uncertainty they cannot capture is assessed in \S\ref{sec:pc_summary}.
The catastrophic-divergence point shows a method-level shift of $+4.5$ epochs in favour of PC, smaller than the PC between-run range ($17$ epochs) by nearly a factor of four.
Neither directional conclusion (``PC is better than SGLD'' or the converse) survives a replication check with two runs per method.

\paragraph{Fixed-epoch comparison.}
Because the margin-10 alignment uses each run's future divergence point and is therefore unavailable prospectively, we repeated the comparison at two common fixed epochs available to all four runs: epoch~90 (before the second scheduled learning-rate decay) and epoch~105 (after it).
The result changes sign between the two epochs: at epoch~90 the method shift is $-0.0089$ macro-5 AUROC (SGLD ahead; exploratory hierarchical $95\%$ CI $[-0.0161,\,-0.0017]$; per-run macro-5: SGLD $0.6871$/$0.6960$ vs PC $0.6801$/$0.6852$), whereas at epoch~105 it is $+0.0182$ (PC ahead; CI $[+0.0031,\,+0.0329]$; SGLD $0.6998$/$0.6827$ vs PC $0.7033$/$0.7155$); the seed-level TOST establishes equivalence at neither epoch ($p = 0.43$ and $0.74$).
Two of the four epoch-105 checkpoints coincide with the margin-10 selection (SGLD \texttt{SEED}~=~42 and PC \texttt{SEED}~=~123), so the two analyses are not independent.
Both effects are comparable to the within-method between-run ranges at the respective epochs, and their sign instability across two adjacent common epochs quantifies the trajectory-phase confound that motivates the margin-10 alignment: at a shared epoch the four runs sit at different distances from their own divergence points ($10$--$27$ epochs at epoch~105), so a fixed-epoch snapshot compares different phases of the trajectories.
The fixed-epoch analysis therefore reinforces the overall conclusion -- no consistent method-level advantage in either direction -- while the margin-10 protocol remains the primary comparison precisely because it controls this phase mismatch.

\subsection{Protocol 2: cold-start generation (FID)}
\label{sec:pc_fresh_fid}

We generate $10\,000$ samples from uniform noise without the replay buffer at an equal computational budget ($1000$ gradient evaluations) on the peak-healthy checkpoint of the canonical SGLD run \texttt{SEED}~=~42 (ep~105, the checkpoint whose buffer-FID is $44.46$), repeating the generation with three independent seeds and pairing the samplers by identical initial-noise streams within each seed.
SGLD reaches FID $52.89 \pm 0.12$ and PC ($K_c = 1$, $500\times (1\mathrm{P}+1\mathrm{C})$) reaches $57.76 \pm 0.26$ (mean $\pm$ sd over seeds); the paired difference $\Delta_{\mathrm{FID}} = +4.87 \pm 0.31$ ($95\%$ paired $t$-interval $[+4.09,\,+5.65]$, $n = 3$) favours SGLD on all three seeds, and KID agrees ($0.0374 \pm 0.0003$ vs $0.0424 \pm 0.0004$).
The $\pm$ dispersion covers generation noise only -- the protocol evaluates a single checkpoint of a single training run, so it does not sample the between-run variability that dominates Protocol~1 (\S\ref{sec:limitations}).
Cold-start generation is thus the one protocol with a consistent detectable difference between the samplers -- in SGLD's favour: at the same gradient budget the adaptation, which replaces half of the noisy Langevin updates with deterministic predictor steps, produces measurably worse cold-start samples.
Whether this reflects the halved stochastic-exploration budget or the inherited parameterisation ($\alpha$, $\sigma$ taken from canonical SGLD) remains untested (\S\ref{sec:limitations}).
Both FID values are markedly worse than the buffer-FID ($44.46$), which agrees with the remark of \citet{Grathwohl2020Your} on the advantage of buffer initialisation.

\subsection{Protocol 3: refinement-style multi-OOD AUROC}
\label{sec:pc_refinement_ood}

The static OOD score $s_{\mathrm{static}}(x) = -f_\theta^{\operatorname{LSE}}(x)$ is sampler-independent; to obtain a sampler-dependent score we use the refinement protocol $s_{\mathrm{refine}}(x_{\mathrm{test}}) = -f_\theta^{\operatorname{LSE}}(x_K)$, where $x_K = \operatorname{Sampler}(x_{\mathrm{test}}, K)$ with $K = 20$ total gradient evaluations (for PC, $K_{\text{outer}} = 10$ with $K_c = 1$).
The evaluation was performed at two qualitatively different checkpoints of the canonical SGLD run \texttt{SEED}~=~42: \textbf{ep~45} (inverted landscape, static AUROC vs SVHN $= 0.4243$) and \textbf{ep~105} (healthy, peak AUROC $= 0.5676$), on five OOD datasets.
Per-image scores yield a paired-bootstrap $95\%$ CI on $\Delta\text{AUROC}_{\mathrm{PC-SGLD}}$ with $B = 2000$ resamples.

\begin{table}[htbp]
    \centering
    \caption{Multi-OOD AUROC: PC ($K_c = 1$) vs SGLD at $K = 20$ total gradient evaluations. $\Delta = \text{AUROC}_{\mathrm{PC}} - \text{AUROC}_{\mathrm{SGLD}}$; $95\%$ CI -- paired bootstrap; $P_{\mathrm{PC}>\mathrm{SGLD}}$ -- estimated probability of a PC advantage.
    The CIs are per-image bootstrap intervals at fixed checkpoints; method-level conclusions rest on the hierarchical seed$\times$image analysis of \S\ref{sec:pc_summary}}
    \label{tab:multi_ood}
    \footnotesize
    \setlength{\tabcolsep}{2pt}
    \begin{tabular}{llccccc}
        \toprule
        \textbf{Epoch} & \textbf{OOD dataset} & \textbf{AUROC} & \textbf{AUROC} & $\boldsymbol{\Delta}$ & \textbf{95\% CI} & $\boldsymbol{P_{\mathrm{PC}>\mathrm{SGLD}}}$\\
        & & \textbf{(SGLD)} & \textbf{(PC)} & & & \\
        \midrule
        \multirow{5}{*}{45 (inv.)}
          & SVHN         & $0.3953$ & $0.3898$ & $-0.0055$ & $[-0.0057,\,-0.0053]$ & ${<}0.0005$ \\
          & CIFAR-100    & $0.4882$ & $0.4885$ & $+0.0003$ & $[-0.0000,\,+0.0006]$ & $0.973$ \\
          & DTD/Textures & $0.6086$ & $0.6082$ & $-0.0004$ & $[-0.0009,\,+0.0001]$ & $0.075$ \\
          & LSUN-R       & $0.6699$ & $0.6766$ & $+0.0068$ & $[+0.0065,\,+0.0070]$ & ${>}0.9995$ \\
          & iSUN         & $0.7012$ & $0.7074$ & $+0.0062$ & $[+0.0059,\,+0.0065]$ & ${>}0.9995$ \\
        \midrule
        \multirow{5}{*}{105 (healthy)}
          & SVHN         & $0.4998$ & $0.4960$ & $-0.0037$ & $[-0.0046,\,-0.0029]$ & ${<}0.0005$ \\
          & CIFAR-100    & $0.5410$ & $0.5420$ & $+0.0010$ & $[+0.0001,\,+0.0018]$ & $0.983$ \\
          & DTD/Textures & $0.6132$ & $0.6119$ & $-0.0012$ & $[-0.0030,\,+0.0006]$ & $0.091$ \\
          & LSUN-R       & $0.5905$ & $0.5939$ & $+0.0034$ & $[+0.0026,\,+0.0043]$ & ${>}0.9995$ \\
          & iSUN         & $0.6060$ & $0.6089$ & $+0.0029$ & $[+0.0020,\,+0.0038]$ & ${>}0.9995$ \\
        \bottomrule
    \end{tabular}
\end{table}

The observed effects are heterogeneous across datasets and small in magnitude.
On the scene-based OOD sets (LSUN-R, iSUN) PC is significantly better than SGLD ($\Delta \in [+0.0029, +0.0068]$); on SVHN it is significantly worse ($\Delta \in [-0.0055, -0.0037]$); on CIFAR-100 and DTD the difference is practically zero (per-image bootstrap at fixed checkpoints; the pattern does not survive the seed-level analysis, \S\ref{sec:pc_summary}).
The direction of the effect depends on the (checkpoint, OOD) pair, with no consistent meta-pattern.
All $|\Delta| < 0.007$, i.e., a practically negligible difference.
\emph{The static score dominates all refinement variants on all $10$ pairs without exception}: at ep~105, SGLD refinement collapses the AUROC vs SVHN from $0.5676$ (static) to $0.4998$ (refined), nearly to chance level.
This conclusion is not an artefact of the inherited step size: across a fourfold range of the PC step ($\alpha \in \{0.5, 1.0, 2.0\}$, SVHN) the PC-refinement AUROC deviates from the $\alpha = 1.0$ setting by at most $0.0130$ ($0.4025 / 0.3898 / 0.3768$ at ep~45; $0.4974 / 0.4960 / 0.5037$ at ep~105) and at no setting approaches the static score ($0.4243$ and $0.5676$).
On the evaluated checkpoints, the best OOD detection on canonical JEM is achieved by \emph{not using a sampler at all}, which further reduces the relevance of the PC-vs-SGLD choice at inference.

\subsection{Summary interpretation and method-level statistical analysis}
\label{sec:pc_summary}

All three protocols yield one consistent picture: \emph{no detectable method-level advantage} of PC over canonical SGLD (Table~\ref{tab:summary}).
At training the cross-method differences are comparable in magnitude to the between-run ranges, at refinement they stay within $|\Delta| < 0.007$ AUROC; the one consistent detectable difference -- cold-start generation quality -- favours SGLD.

\begin{table}[tbp]
    \centering
    \caption{Summary of PC vs SGLD across the three protocols.
    None of the three reveals a method-level advantage of PC over SGLD: at training the differences are comparable in magnitude to the between-run ranges, at refinement they stay within $|\Delta| < 0.007$ AUROC, and in cold-start generation the one consistent detectable difference favours SGLD.
    Refinement AUROC values are means over the five OOD datasets.
    Where two values are given they are the two runs of that method, followed by their mean; the last column carries every cross-method quantity, so no entry belongs to one method alone}
    \label{tab:summary}
    \footnotesize
    \setlength{\tabcolsep}{4pt}
    \begin{tabular}{@{}>{\raggedright\arraybackslash}p{2.9cm}>{\raggedright\arraybackslash}p{2.6cm}>{\raggedright\arraybackslash}p{2.6cm}>{\raggedright\arraybackslash}p{3.4cm}@{}}
        \toprule
        \textbf{Quantity} & \textbf{Canonical SGLD} & \textbf{PC} & \textbf{PC vs SGLD} \\
        \midrule
        \multicolumn{4}{@{}l}{\emph{Protocol~2 -- cold-start FID (inference)}} \\
        FID, 3 paired seeds & $52.89 \pm 0.12$ & $57.76 \pm 0.26$ & $\Delta = +4.87 \pm 0.31$; SGLD better \\
        KID, 3 paired seeds & $0.0374 \pm 0.0003$ & $0.0424 \pm 0.0004$ & SGLD better \\
        \midrule
        \multicolumn{4}{@{}l}{\emph{Protocol~3 -- refinement-style multi-OOD AUROC (inference)}} \\
        Refinement AUROC, ep~45 & $0.5727$ & $0.5741$ & $\Delta = +0.0015$ \\
        Refinement AUROC, ep~105 & $0.5701$ & $0.5706$ & $\Delta = +0.0005$ \\
        Max $|\Delta\text{AUROC}|$, 10 pairs & --- & --- & $0.0068$ (ep~45, LSUN-R) \\
        \midrule
        \multicolumn{4}{@{}l}{\emph{Protocol~1 -- PC instead of SGLD for the full trajectories (training)}} \\
        Divergence epoch (H.3) & ep~116 / ep~122 (mean $119$, range $6$) & ep~132 / ep~115 (mean $123.5$, range $17$) & shift $+4.5$ epochs \\
        macro-5 OOD AUROC (margin-10) & $0.6998$ / $0.7047$ (mean $0.7022$) & $0.7003$ / $0.7155$ (mean $0.7079$) & $\Delta = +0.0056$; $\sim 1.1{\times}$ the SGLD range, $\sim 0.4{\times}$ the PC range \\
        Per-image direction, 4 fixed-checkpoint pairs & --- & --- & 2 favour PC, 1 favours SGLD, and 1 shows no detectable difference \\
        \bottomrule
    \end{tabular}
\end{table}

\paragraph{Hierarchical bootstrap and TOST.}
The seed-level evidence consists of four macro-5 values: $0.6998$ and $0.7047$ for the SGLD runs against $0.7003$ and $0.7155$ for the PC runs, so the method-mean shift is $+0.0056$ while the within-method ranges are $0.0049$ (SGLD) and $0.0152$ (PC), and the four cross-method pairings span $-0.0044$ to $+0.0157$.
For an estimate of the method-level shift that carries this between-run dispersion we ran a hierarchical (seed$\times$image) bootstrap ($B = 3000$): at each iteration we first resample with replacement $2$ runs out of the $2$ available within each method, and then, within each selected run, perform a paired resample of the per-image OOD scores.
With only two runs per method the seed level of this resample rests on minimal support, so we label the resulting interval \emph{exploratory}: the method-level $95\%$ CI on $\Delta_{\text{macro-5}}^{\text{PC-SGLD}}$ is $[-0.0046,\,+0.0161]$ at a point estimate of $+0.0056$ and \emph{contains zero}, i.e., no method-level advantage of either method is detected on the available data.
A sensitivity check confirms that this interval is driven almost entirely by the seed level: freezing the seeds and resampling images alone shrinks the interval to $[+0.0040,\,+0.0073]$ (which would misleadingly exclude zero), whereas resampling seeds alone with the per-run values held fixed reproduces the hierarchical interval almost exactly ($[-0.0044,\,+0.0157]$).
With two runs per method the seed-level percentile interval in fact coincides with the full support of the nine possible cross-method pairings, so it should be read as a range of achievable estimates rather than a coverage-calibrated interval.
Per-image intervals at fixed checkpoints must therefore not be read as method-level evidence.
In the per-OOD breakdown, a significant shift appears only on DTD ($\Delta = +0.0164$, CI $[+0.0017,\,+0.0310]$, in favour of PC, unadjusted for multiplicity across the five datasets); on SVHN, CIFAR-100, LSUN-R, and iSUN no statistically significant method-level difference is found at $\alpha = 0.05$.
In particular, the directional pattern of Table~\ref{tab:multi_ood} (LSUN-R/iSUN in favour of PC, SVHN in favour of SGLD), which appeared stable under the per-image estimates, does not survive the hierarchical CI.

Independently of this, we performed a formal equivalence test: a seed-level Welch TOST at the $\pm 0.01$ AUROC margin yields $p_{\text{TOST}} = 0.33$ (df $\approx 1.2$) and does not reject non-equivalence at $\alpha = 0.05$.
This does \emph{not} mean that the methods are non-equivalent -- it means only that a design with $n = 2$ runs per method is underpowered for a formal equivalence test.
Widening the margin does not rescue the design: at the observed between-run dispersion, equivalence at $\alpha = 0.05$ would first be declared at a margin of $\pm 0.043$ AUROC, far beyond any practically meaningful threshold.
The data are consistent both with the equivalence hypothesis and with the hypothesis of a small directional advantage $|\Delta| \leq 0.0161$ AUROC; formally distinguishing these interpretations requires a larger sample of independent runs (\S\ref{sec:limitations} quantifies the required scale).

\section{Discussion}
\label{sec:discussion}

\subsection{Why the fixed-noise adaptation is not expected to improve canonical JEM}
\label{sec:why_pc_fails}

The theoretical advantage of the PC sampler \citep{song2021scorebased} over plain Langevin dynamics arises not from the ``predictor $+$ corrector'' structure itself but from the specific role this structure plays within an \emph{annealed} stochastic differential equation.
Canonical PC is designed to integrate the reverse-time SDE along a noise schedule, where for the VE variant the predictor step takes the explicit form $x'_{i} \leftarrow x_{i+1} + (\sigma_{i+1}^{2} - \sigma_{i}^{2})\,s_{\theta}(x_{i+1},\,\sigma_{i+1})$, with the step length \emph{directly} proportional to the difference $(\sigma_{i+1}^2 - \sigma_i^2)$ between adjacent noise levels.
It is precisely this coupling that guarantees that, in combination with the corrector locally keeping the trajectory on the target density, the resulting trajectory converges to a sample from $p_0(x)$.

For canonical JEM this coupling is broken.
The noise is fixed ($\sigma = 0.01$), the difference between adjacent noise levels is identically zero, and the canonical VE predictor degenerates by construction -- its step vanishes.
In our adaptation (\S\ref{sec:pc_impl}) we replace this degenerate predictor with a simple deterministic gradient step, which is in no way different from an SGLD-like recurrence.
Moreover, JEM has no continuum of marginals $p_t$ but a single static target density, so the predictor in our adaptation cannot perform the cross-marginal function it was designed for.
By the same token, this argument explains the absence of a detectable method-level difference at training time in Protocol~1: at the same budget of $40$ gradient evaluations, PC and SGLD have a matched first-order deterministic-drift budget and differ only in the noise budget ($20\sigma^2$ for PC vs $40\sigma^2$ for SGLD per iteration).
At a small $\sigma = 0.01$, and with a replay buffer whose stationary variance adapts to the total noise budget, this difference is quantitatively small and does not materialise as a detectable method-level effect.
The Protocol-2 asymmetry fits the same picture from the other side: in cold-start generation, where mixing from uniform noise relies on stochastic exploration alone (no replay buffer), the halved noise budget is a plausible -- though untested -- explanation for the consistent FID/KID deficit of the adaptation (\S\ref{sec:pc_fresh_fid}).
Independent confirmation comes from the asymptotics: as $K_c$ grows, the share of predictor steps in the budget shrinks, and $|\Delta\text{AUROC}|$ tends monotonically to zero (SVHN at ep~105: $0.0037 \to 0.0018 \to 0.0004$ for $K_c \in \{1, 3, 5\}$).

\subsection{Structural solutions rather than sampler-level ones}
\label{sec:structural_solutions}

Replacing the sampler within the SGLD family yields no statistically detectable gain on canonical JEM, and Mode~1 occurs in all four runs in the canonical setting with an identical signature -- a limitation that appears structural within the tested canonical configuration (a single architecture, dataset, and hyperparameter set) rather than an artefact of the sampler.
Eliminating the failure modes is likely to require changes that go beyond the static noise regime.
Two directions already carry empirical support in the literature: (i)~bounded deterministic samplers (PGD in DAT \citep{yin2025joint}, FID $\approx 9$ on CIFAR-10) and (ii)~EBM-diffusion hybrids with a native noise schedule \citep{geng2024improving, guo2023egc, compositioncontrol2025}, which restore the annealed-noise setting in which the PC guarantees apply; a third, (iii)~cooperative training \citep{zhu2024learning} (CDRL), where the EBM, the diffusion model, and the generator stabilise one another, we list as a design direction rather than a tested remedy for the failure modes documented here.
All three directions depart from the formulation ``canonical JEM with an alternative sampler''.

A fourth consideration is infrastructural rather than algorithmic, and it follows from the cost figures of this study.
Canonical JEM training is long -- a single run occupies an A100 for $26.9$--$30.8$~h, the divergence arrives only after more than a hundred epochs of otherwise healthy training, and the adequately powered comparison of \S\ref{sec:limitations} would require some $680$ A100-hours -- so the granularity of checkpointing and of diagnostic logging is itself a design variable rather than an implementation detail.
Our five-epoch checkpoint stride preserved usable states on either side of every divergence, but the per-batch records that would identify the triggering sample were not retained, and in two of the four runs the aggregate metrics of the collapse epoch were lost to the mass abort (\S\ref{sec:fm1}).
Training infrastructure that keeps fine-grained diagnostics under a rare-event trigger and can restart automatically from the last healthy state would not remove the failure mode, but it would make each occurrence informative rather than merely expensive.
The two halves of that requirement differ sharply in cost: the diagnostics that were lost here are scalar summaries, negligible beside the model state, whereas retaining full checkpoints at a comparable cadence would dominate the storage and input/output budget of a run.
A trigger that raises the retention rate only around a detected anomaly is therefore the part of the design that carries the engineering burden, and the epoch-level stride that is affordable for a whole run is precisely what loses the batch on which the diagnosis depends.
Our own runs write their checkpoints to a local disk, but at the scale on which such a study would have to be repeated that retention moves onto shared or distributed storage, where its throughput and recovery cost become a systems problem in their own right.
Hardware acceleration of erasure-coded recovery has been pursued at exactly that layer, both in hybrid CPU-FPGA designs \citep{lei2026hcfr} and in elastic multi-block recovery for Ceph \citep{lei2024symmetrical}.
For training regimes of this cost and fragility, the co-design of the model with the hardware and software it runs on is therefore a legitimate part of the structural picture, alongside the three algorithmic directions above.

\subsection{Limitations}
\label{sec:limitations}

\begin{itemize}
    \item \textbf{A design with two runs per method.} Our finding that no method-level PC advantage is detected rests on $n = 2$; this is the first paired replication test of PC vs SGLD on canonical JEM, but it neither yields a narrow CI on the method shift nor establishes formal equivalence at the $\pm 0.01$ AUROC margin (TOST $p = 0.33$), and the retrospective power of the design is effectively zero.
    A prospective power calculation quantifies the required scale: at the observed between-run dispersion (pooled per-run standard deviation $\approx 0.008$ macro-5 AUROC) and assuming a true method difference of zero, a seed-level Welch TOST at the $\pm 0.01$ margin with $80\%$ power at $\alpha = 0.05$ needs roughly $12$ runs per method ($15$ for $90\%$ power; $45$ per method at a $\pm 0.005$ margin) -- at the measured $\approx 28$~h per run, about $680$ A100-hours for a single sampler comparison, an order of magnitude beyond typical replication budgets in EBM research.
    Because the dispersion estimate itself rests on $\hat{\nu} \approx 1.2$ effective degrees of freedom, these counts are order-of-magnitude guidance rather than certified sample sizes; a two-stage design (a pilot of $4$--$6$ runs to stabilise the variance estimate, then a recomputed $n$) is the sound path for future work.
    \item \textbf{The PC between-run range is three times wider than that of SGLD} ($0.0152$ vs $0.0049$ AUROC; $17$ vs $6$ epochs of the divergence-point range).
    This points to a greater seed sensitivity of PC -- for practical use, an additional cost not compensated by a statistically detectable method-mean gain -- although two runs per method cannot establish this contrast as systematic.
    \item \textbf{Protocols 2 and 3 are single-model evaluations.} Cold-start generation and refinement OOD scoring use checkpoints of one training run (\texttt{SEED}~=~42 in the canonical setting), and the Protocol-2 dispersion covers generation seeds only; neither protocol samples the between-run variability that \S\ref{sec:pc_summary} shows to dominate the training comparison, so their conclusions are conditional on the evaluated model.
    \item \textbf{The margin-10 checkpoint-selection protocol is a post-hoc alignment.} Checkpoints are matched by the distance to each run's own catastrophic-divergence point; this avoids the unfair comparison of one method's ``best of all epochs'' against a fixed epoch of the other, but it is a post-hoc normalisation, sensitive to the accuracy of identifying the divergence epoch ($\pm 1$--$2$ epochs given the checkpointing stride of $5$).
    \item \textbf{A restricted PC regime and parameterisation.} Only PC with fixed $\alpha$ and $\sigma$ was tested ($K_c \in \{1, 3, 5\}$ at inference, $K_c = 1$ for the training-time replacement); PC with an annealed $\sigma_t$ (the only variant that theoretically restores the guarantees of \citet{song2021scorebased}) takes us into the space of EBM-diffusion hybrids.
    Moreover, the training-time hyperparameters of the adaptation ($\alpha = 1.0$, $\sigma = 0.01$) are inherited from canonical SGLD rather than tuned for the adapted kernel, so the training-protocol conclusion is, strictly, that \emph{the selected PC parameterisation} provides no detectable advantage; an inference-level step-size check (\S\ref{sec:pc_refinement_ood}) shows the refinement conclusion is stable across a fourfold $\alpha$ range, but training-time tuning remains untested.
    \item \textbf{A single hyperparameter set and a single base setting.} Canonical parameters \citep{Grathwohl2020Your}, WideResNet-28-10, CIFAR-10.
    Generalisation to other JEM variants (SADA-JEM \citep{yang2023towards}, \citep{jiang2025your}), architectures, and datasets (ImageNet) remains for future work; the theoretical argument of \S\ref{sec:why_pc_fails} retains its force as long as the noise remains fixed.
    \item \textbf{A low-AUROC regime.} Canonical JEM is itself a weak OOD detector in this setting (best static AUROC vs SVHN $0.608$ against the canonical reference $0.670$); the no-detectable-advantage verdict is therefore obtained in a low-AUROC regime and need not transfer to substantially stronger detectors.
\end{itemize}

\section{Conclusions}
\label{sec:conclusions}

We performed a high-fidelity reconstruction of the canonical JEM \citep{Grathwohl2020Your} on WideResNet-28-10 with no normalisation layers on two independent runs; it reproduces the canonical classification accuracy to within $0.02$~pp, with a residual buffer-FID gap of $+6.06$ that may partly reflect the limited training horizon.
Two failure modes are documented: catastrophic divergence with the Appendix~H.3 signature, identical in all four runs in the canonical setting, and run-dependent SVHN OOD-discrimination dynamics (\texttt{SEED}~=~42 oscillates, \texttt{SEED}~=~123 improves steadily over epochs~100--120).

On the three protocols testing the PC hypothesis, no method-level advantage of PC over canonical SGLD is detected: at refinement, $|\Delta\text{AUROC}| < 0.007$ for all $10$ (ep, OOD) pairs; in seeded cold-start generation the one consistent detectable difference favours SGLD ($\Delta_{\mathrm{FID}} = +4.87 \pm 0.31$ at the matched budget); on the training protocol, the hierarchical (seed$\times$image) bootstrap ($B = 3000$) yields a $95\%$ CI on $\Delta_{\text{macro-5}} \in [-0.0046,\,+0.0161]$ that contains zero, and a Welch TOST at the $\pm 0.01$ AUROC margin with $n = 2$ runs per method does not establish formal equivalence ($p = 0.33$).
The non-detection of a method-level advantage is theoretically consistent with the limitation of the PC framework of \citet{song2021scorebased}: under fixed $\sigma$ the predictor step degenerates by construction, so the theoretical guarantees do not transfer to canonical JEM.

The practical implications within the tested setting are as follows: (1)~among interventions at the level of the SGLD sampler family itself, no practical gain from PC is detected on the available evidence -- in cold-start generation the adaptation is in fact detectably worse -- while formal equivalence on the training protocol is not established either; (2)~the choice of the inference-time sampler is secondary -- the static score $-\operatorname{LSE} f_\theta$ dominates all refinement variants on all $10$ tested pairs; (3)~eliminating the documented failure modes is likely to require structural changes outside the static noise regime (PGD in DAT \citep{yin2025joint}, EBM-diffusion hybrids \citep{geng2024improving, guo2023egc}, cooperative training \citep{zhu2024learning}); (4)~the $n = 2$ vs $n = 2$ design with a hierarchical seed$\times$image bootstrap, a seed-level TOST, and a divergence-distance-matched checkpoint-selection protocol constitutes a minimal statistical standard for further sampler comparisons on canonical JEM.

\backmatter

\bmhead{Reproducibility}
All four runs in the canonical setting were executed on a single NVIDIA A100 GPU (40~GB) in the Modal Cloud environment; the stack is PyTorch 2.x, Python 3.11.
Because individual training jobs are subject to a 24-hour wall-clock limit, longer runs were checkpointed and resumed from the latest checkpoint with the optimiser and replay-buffer state restored, so that resumption continues the same trajectory rather than constituting an independent run.
At the matched budget of $40$ gradient evaluations per batch, the mean epoch duration is $855$/$853$~s for the two SGLD runs and $846$/$849$~s for the two PC runs (method means $854.0$ vs $847.5$~s; $26.9$--$30.8$~h per full run), so the adapted sampler carries no wall-clock overhead -- it is $\approx 0.8\%$ cheaper in both runs of each method ($n = 2$), drawing half as many Gaussian noise tensors per batch.
At inference, one seeded Protocol-2 generation pass ($10^4$ samples, both samplers, FID/KID included) takes $\approx 2$~h on the same GPU, and Protocol-3 refinement scoring of all six datasets completes within minutes per checkpoint.
Peak GPU memory of the training step, measured by a $10$-batch instrumented probe on the epoch-105 checkpoint, is identical for the two samplers ($4.96$~GiB allocated for both).
The samplers do differ in the number of stochastic noise injections at the same budget: $40$ (SGLD) vs $20$ (PC) per training batch in Protocol~1, $1000$ vs $500$ per sample in Protocol~2, and $20$ vs $10$ per test image in Protocol~3.
Buffer-FID is computed by the canonical protocol \citep{Grathwohl2020Your}: $10\,000$ samples from the buffer without additional SGLD steps.
Paired AUROC bootstrap: joint resampling of the in- and out-distribution indices with replacement, with the AUROC of both methods computed on the same resampled indices; $B = 2000$ for the per-image Protocol-3 CIs and $B = 3000$ for the margin-10 comparisons of Figure~\ref{fig:pc_vs_sgld}, where the macro-5 statistic is resampled jointly (one shared in-distribution index vector and independent per-dataset out-distribution vectors per iteration), so the interval applies to the macro average itself.
The full hyperparameter specification is given in \S\ref{sec:canonical}; the experiment-reconstruction code, the OOD/FID evaluators, the bootstrap analyses, additional diagnostic tables, and the per-image scores for all four runs in the canonical setting are published in the project repository: \url{https://github.com/DmytroKnopov/jem-sampler-comparison}, archived at \url{https://doi.org/10.5281/zenodo.21872469}.

\bmhead{Acknowledgements}

The author dedicates this publication to the cherished memory of his academic advisor, Associate Professor Dr.~Ruslan Chornei, and expresses deep gratitude for his guidance, support, and inspiration along the research path.

\medskip
\noindent
This work was carried out with the support of the project \emph{``Platform for Model Based Solutions to Systemic Complexity Challenges: Bridging Leading Research Schools of Finland and Ukraine by Double Doctor Degree''}.

\bmhead{Use of AI-assisted tools}

During the preparation of this manuscript, the author used large language models from Anthropic (Claude Opus) and Google (Gemini Pro) to assist with drafting and editing the text, organising the related literature, and developing and debugging the code for the experiments.
The research ideas, methodology, experimental design, analyses, results, and conclusions reported here are entirely the author's own original work, for which the author takes full responsibility.

\section*{Declarations}

\bmhead{Funding}
The author received no specific funding for this work.

\bmhead{Competing interests}
The author declares no competing interests.

\bmhead{Ethical approval}
Not applicable.
This study involves no human participants and no animals; it uses only publicly available benchmark image datasets, so approval by an ethics committee was not required.

\bmhead{Consent to participate}
Not applicable.
No human participants were involved in this study.

\bmhead{Consent to publish}
Not applicable.
The manuscript contains no data, images, or videos relating to an identifiable individual person.

\bmhead{Clinical trial number}
Not applicable.

\bmhead{Data availability}
The datasets generated during and/or analysed during the current study (per-image out-of-distribution scores, FID measurements, and bootstrap outputs for all four runs in the canonical setting) are available in the Zenodo repository, \url{https://doi.org/10.5281/zenodo.21872469}.
The third-party benchmark datasets analysed in this study (CIFAR-10, SVHN, CIFAR-100, DTD/Textures, LSUN-R, iSUN) are publicly available.

\bmhead{Materials availability}
Not applicable.

\bmhead{Code availability}
The experiment-reconstruction code, the out-of-distribution and FID evaluators, and the bootstrap analyses are available in the project repository, \url{https://github.com/DmytroKnopov/jem-sampler-comparison}, archived at \url{https://doi.org/10.5281/zenodo.21872469}.

\bmhead{Author contributions}
Dmytro Knopov is the sole author.

\bibliography{references}

\end{document}